\documentclass[a4paper, 10pt, conference]{ieeeconf}      

\IEEEoverridecommandlockouts                              
\usepackage{amsmath}
\usepackage{amssymb}
\usepackage{amsfonts}
\usepackage[pdftex]{graphicx}
\DeclareGraphicsExtensions{.pdf, .png, .jpg}
\usepackage{algorithm}
\usepackage[noend]{algpseudocode}
\algrenewcommand\algorithmicindent{0.7em}%
\usepackage{gensymb}
\usepackage[caption=false]{subfig}
\usepackage{cite}
\usepackage{hyperref}
\usepackage{flushend}
\usepackage{balance}
\usepackage{algpseudocode}

\usepackage{amsthm}

\theoremstyle{definition}

\usepackage[usenames,dvipsnames]{xcolor}

\title{\LARGE \bf
Decentralised Intelligence, Surveillance, and Reconnaissance in Unknown Environments with Heterogeneous Multi-Robot Systems
}

\author{Ki Myung Brian Lee$^1$, Felix Kong$^1$, Ricardo Cannizzaro$^2$, Jennifer L. Palmer$^2$, \\
David Johnson$^3$, Chanyeol Yoo$^1$ and Robert Fitch$^1$ 
\thanks{This work is supported by an Australian Government Research Training Program (RTP) Scholarship, Australia's Defence Science and Technology Group, and the University of Technology Sydney.}
\thanks{$^1$Authors are with the University of Technology Sydney, Ultimo, NSW 2006, Australia {\tt\footnotesize brian.lee@student.uts.edu.au, \{chanyeol.yoo, felix.kong, rfitch\}@uts.edu.au}}
\thanks{$^2$Authors are with the Defence Science and Technology Group, Department of Defence, Australia {\tt\footnotesize \{ricardo.cannizzaro, jennifer.palmer\}@dst.defence.gov.au}}
\thanks{$^3$Author is with Mission Systems Pty. Ltd., Sydney, Australia {\tt\footnotesize david.johnson@missionsystems.com.au}}
}

\begin{document}

\maketitle

\begin{abstract}
We present the design and implementation of a decentralised, heterogeneous multi-robot system for performing intelligence, surveillance and reconnaissance~(ISR) in an unknown environment.
The team consists of functionally specialised robots that gather information and others that perform a mission-specific task, and is coordinated to achieve simultaneous exploration and exploitation in the unknown environment.
We present a practical implementation of such a system, including decentralised inter-robot localisation, mapping, data fusion and coordination.
The system is demonstrated in an efficient distributed simulation.
We also describe an UAS platform for hardware experiments, and the ongoing progress. 
\end{abstract}

\section{Introduction}
There is an increasing demand for robotic systems to explore and operate in unknown, possibly adverse environments, for applications such as disaster relief~\cite{search_and_rescue_robots} and defence~\cite{cooperative_air_and_ground}. 
The unknown and adverse nature of the environment is the very reason that robotic systems are useful, because it frees the human operators from the potential hazards.
Simultaneously, these environmental properties lead to an important fundamental challenge in robotics, namely, to ensure reliable, robust operation of robots against uncertainty.
Decentralised multi-robot systems are a prominent approach to ensuring reliability and robustness in practice, because many inexpensive, possibly disposable robots can be composed into an effective team while ensuring redundancy.

Heterogeneity allows further improvements via functional specialization; different robots can be equipped with different payloads to enhance the team's overall capability.  
To this end, we recently proposed the concept of \emph{`scout-task architecture'}~\cite{brian_MIUCB_2021}, where \emph{scout} robots are equipped with sensors to explore the environment and gather information, while \emph{task} robots are equipped with a payload to execute the desired task. 
We showed that a scout-task composition can side-step the fundamental challenge of exploration-exploitation trade-off in decision-making under uncertainty by performing both simultaneously.

\begin{figure}[t!]
    \centering
    \includegraphics[width=0.97\columnwidth]{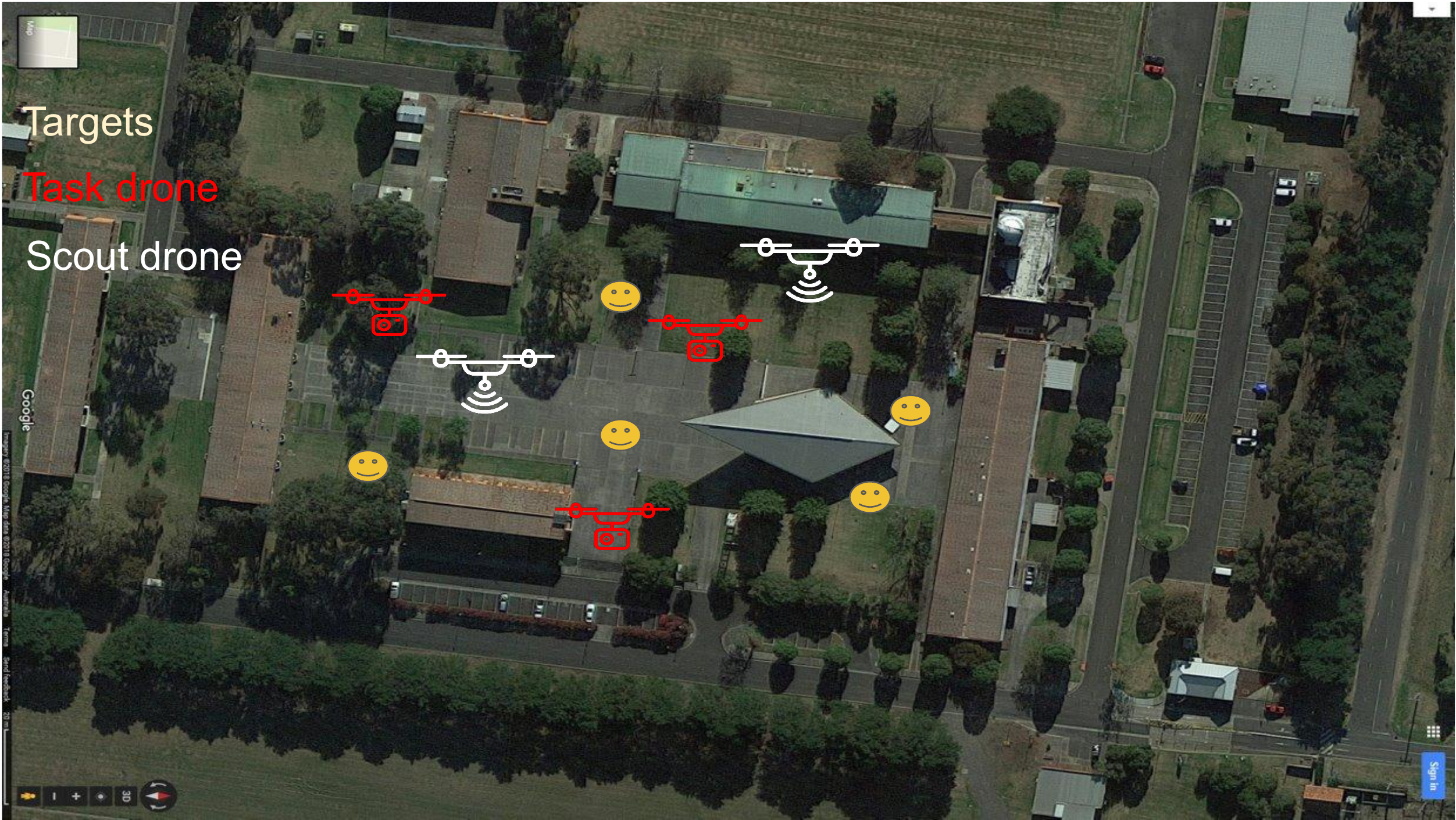}
    \caption{We consider multi-drone surveillance in an unknown environment. The task is to find and detect all targets (yellow) with the task drones (red), which carry short-range sensors that make high-confidence measurements. Scout drones (white) support the task drones by sensing targets from a distance using their long-range sensors, which low-confidence measurements, and communicate potential target locations to the task drones. Some drones are both scout and task drones. \label{fig:scenario}}
\end{figure}

In this paper, we present the system architecture for realizing scout-task coordination for an intelligence, surveillance and reconnaissance~(ISR) task considered in~\cite{brian_MIUCB_2021}.
The goal is to visually confirm all targets in the environment for verification by human operators (see Fig.~\ref{fig:scenario}), which is useful in disaster relief and military scenarios. 
The \emph{task robots} are equipped with short-range visual sensors that achieve the goal directly, and the \emph{scout robots} are additionally equipped with long-range sensors that identify possible presence of targets. 
The environment is initially unknown, and no global localisation source~(e.g. GPS) is available.

To this end, we develop a decentralised software framework that consists of inter-robot localisation, mapping and planning based on the robot operating system~(ROS)~\cite{ros}.
The inter-robot localisation and mapping module extends the popularly used real-time appearance based mapping~(RTABMap) library \cite{labbe_rtabmap} to \textit{online} multi-robot simultaneous localisation and mapping (SLAM).
The planning module realises the mutual information upper confidence bound (MI-UCB) strategy~\cite{brian_MIUCB_2021} using decentralised Monte Carlo tree search~(Dec-MCTS)~\cite{decmcts}.  
We demonstrate the proposed framework in a distributed high-fidelity simulation, and discuss the ongoing progress in hardware experiments. 

\section{Related Work}

Multi-robot ISR is typically posed as a search problem~\cite{hollinger_survey}, where the goal is to coordinate a team of robots to find targets in the environment.
Sensor measurements are fused into a probabilistic belief over target locations, represented as an occupancy grid~\cite{vidal_pursuit_evasion} or a random finite set~ \cite{clark2006gm,dames,Sung2018}.
Subsequently, the robots are coordinated to maximise the expected number or probability of detections \cite{pursuit_evasion_uav_ugv,optimal_search_for,hollinger_multi_robot_moving,hollinger_survey,seng_keat_gan_collision} or to minimise the uncertainty of target locations \cite{brent_schotfeldt,cliff2018robotic,dames}.
To this end, previous approaches considered heterogeneity in terms of mobility~\cite{pursuit_evasion_uav_ugv,cooperative_air_and_ground,rangers_and_scouts} or sensing capability~\cite{smith_and_hollinger,aerial_heterogeneous}.
Our notion of `scout-task composition' generalise the scope of heterogeneity to mission-level capability. 
While a set of robots complete the task at hand directly, another set of robots assist implicitly by gathering environmental information, as in the case of the Perseverance Mars rover and the Ingenuity helicopter~\cite{ali_agha_mohamadi_where_to_map,mohammadi_2}.
In the context of ISR, this allows greater control over the information gathered by the team. 

\section{Scenario Description}
The robot team consists of UAV drones that belong to two classes - RGBD and LIDAR.
RGBD drones are \emph{task-only robots} that provide visual confirmation which is the intended task.
LIDAR drones are \emph{scout-and-task robots}, and generate a set of detections of the targets over a long-range, omnidirectional field of view~(FOV).
LIDAR drones are also equipped with RGBD sensors, and hence also provide visual confirmation if possible.
The use of LIDAR sensors allows faster exploration of the unknown environment so that possible target locations can be communicated to teammates. 

\section{Sensors and Software Architecture} \label{sec:software}
\begin{figure}
    \centering
    \includegraphics[width = \columnwidth]{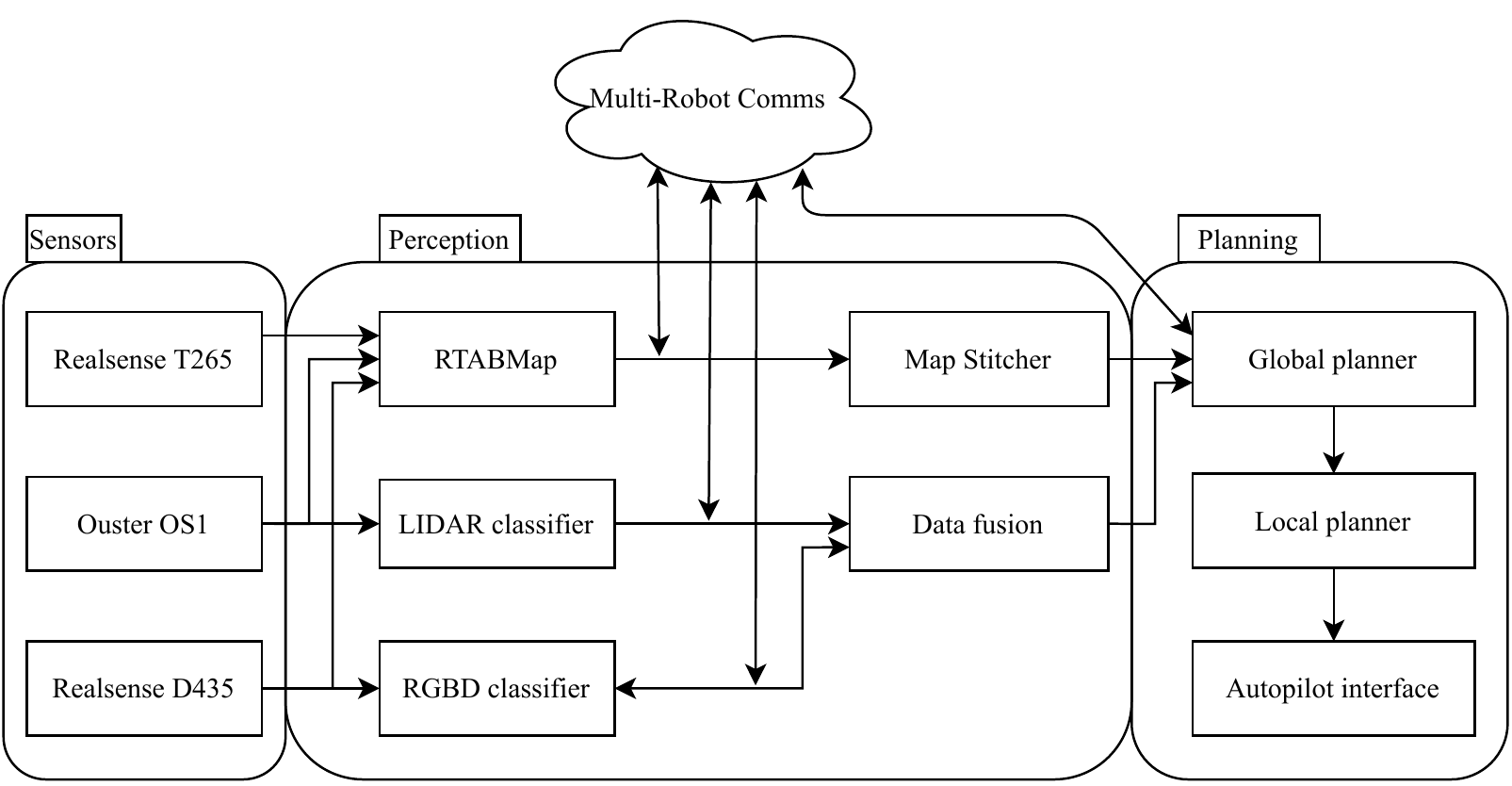}
    \caption{Software system overview for a single LIDAR drone. RGBD drones do not have the LIDAR sensor onboard.}
    \label{fig:system_architecture}
\end{figure}

In this section, we present the sensors and software components of the framework, as summarised in Fig.~\ref{fig:system_architecture}.


\subsection{Sensors}\label{sec:sensors}
The sensor payload for the RGBD drones consists of an Intel RealSense D435 RGBD camera mounted facing forward for target recognition, and a nadir-pointing Intel RealSense T265 RGBD camera for localisation and mapping.
The T265 camera also includes an onboard simultaneous localisation and mapping (SLAM) implementation, and can used to produce a pose estimate of a drone relative to its own initial pose.
The T265 camera SLAM solution is considered a source of odometry in a higher-level SLAM implementation (see Section \ref{sec:mappingAndLocalization}).

The D435 camera also contributes towards mapping and localisation; the RGBD images it takes are fed to the aforementioned higher-level SLAM implementation. 
In addition, the the RGBD images are also used as an input to a deep image classifier to recognise targets.
Target detections from the RGB classifier are considered to be true detections.
In ideal indoor conditions, the D435 is rated to give accurate depth readings at ranges up to 10m, though in our outdoor experiments, this was never realised.

The LIDAR drones also carry T265 and D435, but is augmented by an Ouster OS1 3D LIDAR. 
The OS1 LIDAR provides an omnidirectional field of view with an order of magnitude longer sensor range than the RealSense D435 (about 100m), allowing the robots carrying the OS1 to sense large areas quickly.
However, a LIDAR scan was considered to have insufficient resolution to reliably decide if a segment of the scan is a target or not; especially at long ranges, the LIDAR point cloud can be very sparse.
The LIDAR sensor data was used as an input to the higher-level SLAM algorithm, as well as to detect \textit{potential} targets using the LIDAR classifier.
Targets identified using the LIDAR classifier are considered potential targets, and require confirmation using the shorter-range D435.

\subsection{Target Detection and Decentralised Fusion}\label{sec:classifiers}
To detect the presence of a target in a sensor measurement, classifiers were trained to segment humanoids from LIDAR and RGBD point clouds. 
The RGBD classifier is based on the \texttt{jetson-inference}\footnote{https://github.com/dusty-nv/jetson-inference} library, which was modified to correlated RGB and depth images from the RealSense D435 camera.
We use the LIDAR classifier proposed in~\cite{lidar_classifier}, which consists of a cluster extraction algorithm, an SVM classifier, and a nearest-neighbour-based Kalman filter for data association and tracking.
The LIDAR classifier required additional training data, which was sourced from some field trials.

The detection results from the RGBD and LIDAR classifiers are broadcast to the entire team.
Each drone runs an independent occupancy grid filter described in~\cite{brian_MIUCB_2021} to fuse the detections generated by itself and the ones communicated by others.  

\subsection{Decentralised mapping and localisation}\label{sec:mappingAndLocalization}
For a drone to perform its mission, it must be able to map its environment and localise itself in the map.
Each drone runs real-time appearance-based mapping (RTABmap) \cite{labbe_rtabmap} onboard using sensor measurements from the Realsense T265, D435, and if available, the Ouster OS1 and global positioning system (GPS) sensor.
RTABmap enables each drone to build a map of its surroundings, assign a geographic location to detected targets, and to perform reliable path-planning and navigation.
The C++ RTABmap package interfaces with ROS via the \texttt{rtabmap\_ros} node, which allows other processes (e.g. the global planner) access to the map and current drone pose estimate generated by RTABmap.

In addition to decentralised operation, another requirement of the system is that it can operate in global navigation satellite system (GNSS) denied scenarios. 
Since the system consists of multiple drones, a common geographic reference frame must be established online between drones without the use of GNSS services.
An online ``map stitching'' ROS node was developed to ``stitch'' maps of different drones together, thereby joining the maps between drones, and thus allowing a common reference frame. 
In particular, the relative 3D rotation and translation between two drones' individual SLAM maps is estimated periodically, facilitated by sharing of downsampled sensor data.
This is the main idea behind the  \texttt{map\_stitcher} node, which simply performs loop closures \textit{between} drones, as opposed to between poses of a single drone. 
While this functionality exists in RTABmap during offline postprocessing, the \texttt{map\_stitcher} node was developed for this project for online map stitching, with modifications for fast execution. 

We present a high-level overview of the map stitching process (see Algorithm \ref{alg:mapStitching}). 
Let $Z_i$ be the set of all of its sensor measurements of the environment.
The elements $z\in Z_i$ are ``full'' sensor measurements, with geometric information about the environment, in contrast to the ``measurements'' $Y$, which refer to target detections by the classifier.
Then, given an initial pose, the internal SLAM algorithm of drone $i$ will produce a pose graph $G_i$, which can be solved to estimate its trajectory $T_i$, i.e. the set of all of its poses. 
Drone $i$ the receives compressed sensor measurements from other drones, and uses them to perform a map stitching.

\begin{algorithm}[t]
    \caption{Map stitching algorithm}\label{alg:mapStitching}
    \begin{algorithmic}[1]
        \Procedure{StitchMap}{$ G_i,G_j,z_j(t)$}
        \For{$z_k\in Z_i$}
            \State $s \gets$ \Call{SimilarityScore}{$z_j,z_k$}
            \If{$s > s^\star$}
                \State $\hat T \gets$\Call{EstimateRelativePose}{$z_j,z_k$} 
                \If{$\hat T \neq \emptyset$}
                    \State \Call{AddLoopClosure}{$G_i,G_j,p_k,p_j,\hat T$}
                \EndIf
            \EndIf
        \EndFor
        \State$\mathcal G \gets$ \Call{SolvePoseGraph}{$G_i\bigcup G_j$}
        \State \textbf{return} $\mathcal G$
        \EndProcedure
    \end{algorithmic}
\end{algorithm}

Suppose drone $i$ receives an  incoming sensor measurement $z_j$ at pose $p_j$ from drone $j$ and its pose graph $G_j$. 
The map stitching algorithm running on drone $i$ compares $z_j$ to all of its own measurements $Z_i$.
The function \Call{SimilarityScore}{} is a fast way to estimate the ``similarity'' between two RGB images; if the images are similar enough, relative pose estimation is attempted.
If the relative pose between the two estimates is successful, then a the relative pose (along with covariance information) is added between the two pose graphs $G_i$ and $G_j$. 
If multiple measurements have high similarity score and successful relative pose estimates, multiple loop closures can be added between $G_i$ and $G_j$ from $z_k$.
When all the measurements in $Z_i$ have been checked against the incoming $z_j$, the joint pose graph is solved.
Hence the maps of drones $i$ and $j$ are joined together in $\mathcal G$. 
This allows a shared geographic reference between the drones $i$ and $j$ which can operate in GNSS-denied scenarios, facilitating the sharing of target beliefs among drones.
\subsection{Global planner}\label{sec:globalPlanner}
Based on the results of data fusion and obstacle mapping, the global planner generates a high-level plan that coordinates the scout-task team.
We use the mutual information upper confidence bound~(MI-UCB) acquisition function proposed in~\cite{brian_MIUCB_2021}.
Intuitively, the MI-UCB is a weighted sum of information gain of the measurements provided by the scout robots, and the prior expected reward\footnote{To be precise, it is the cumulant generating function~(CGF) of the prior expected reward} before scout robots provide measurements:
\begin{equation}\label{eq:mi_ucb}
    U_{\text{MI-UCB}} \triangleq \frac{1}{\delta} I(E; Y | A) + \log \mathbb{E} \exp R(E, A).
\end{equation}
Here, $E$ represents the belief over target presence, as computed by the occupancy grid filter, $A$ represents the set of trajectories executed by the team, and $Y$ represents the detection measurements gathered by the scout robots. 
$ I(E; Y | A) $ is the Shannon information between $E$ and $Y$ given $A$ is executed, and $R(E, A)$ is the reward~(number of targets visually confirmed).

The central result of~\cite{brian_MIUCB_2021} is that~\eqref{eq:mi_ucb} is an upper bound on the \emph{posterior} expected reward given measurements with probability $\geq 1 - \delta$:
\begin{equation}
    \mathbb{E}[ R(E, A) | Y] \leq U_{\text{MI-UCB}}.
\end{equation}

This relationship allows us to maximise for the posterior expected reward $\mathbb{E}[ R(E, A) | Y]$ without explicit knowledge of the measurements $Y$ in a no-regret sense.
In practical terms, the exploratory behaviour induced by the Shannon information term causes the team to simultaneously visually confirm the identified potential targets (i.e. exploit) and search for other potential targets (i.e. explore).
For example, if there is a target near a LIDAR drone and a RGBD drone, the resulting behaviour that maximises~\eqref{eq:mi_ucb} is for RGBD drone to undertake the visual confirmation and the LIDAR drone to continue exploring. 

The MI-UCB acquisition function is maximised using the Dec-MCTS algorithm~\cite{decmcts}. 
Briefly, Dec-MCTS is a decentralised algorithm that allows the team of drones generate a set of trajectories $A$ that maximises the acquisition function~\eqref{eq:mi_ucb}.
In Dec-MCTS, each robot maintains a probability distribution over the trajectories of the entire team.
The distribution over other robots' trajectories are updated through communication~(Sec.~\ref{sec:interfaces}).
Each robot updates the distribution over its own trajectory through single-robot MCTS iterations, during which sampled instances of the other robots' trajectories are used to evaluate and maximise the acquisition function~\eqref{eq:mi_ucb} evaluated for the whole team.
\subsection{Local planner}\label{sec:localPlanner}
We use the local obstacle avoidance planner proposed in~\cite{3dvfh}.
Its role is to take the planned paths from the global planner and generate a local collision-free trajectory near the drone's current location for the low-level PX4Cube to follow.

The local planner uses the last few frames from the D435 to map the obstacles. 
This ensures two layers of redundancy against unmapped or moving obstacles. 
The global plan is collision-free with respect to the large-scale map built by the decentralised mapping module, and the local plan is collision-free with respect to the local RGBD measurements, which does not depend on the decentralised mapping module. 

\section{Communication Architecture} \label{sec:interfaces}

\subsection{ROS and MOOS networks}
Robot Operating System (ROS) \cite{ros} is a software library that provides a common interface for different C++ and Python programs, called \textit{nodes}, to communicate with one another. 
ROS is used onboard each drone to manage communications within the drone's NVIDIA Jetson TX2 computer.
However, ROS cannot be used in a decentralised way between multiple computers, as it requires a unique ROS ``master'' to be run on the network.
The failure of this ROS ``master'' would bring down the entire team, which is undesirable.
Hence, a different middleware solution called Mission Oriented Operating Suite (MOOS) \cite{moos} is used for inter-vehicle communications. 
MOOS has the advantage that it can be used in a decentralised way, without a single ``master'' computer.
With most of the algorithms being written in ROS, including the local and global planners, the classifiers, the SLAM algorithm, and the sensor drivers, bridge between ROS and MOOS was used, called \texttt{rospymoos}\footnote{https://github.com/SyllogismRXS/moos-ros-bridge}. 
We developed a modified version of \texttt{rospymoos} to handle custom ROS messages required in this project.
Any ROS messages containing information necessary to be sent offboard the drone are converted into a MOOS message, and sent to other drones over WiFi via MOOS. 
Similarly, messages from other drones are received by MOOS and converted into ROS-compatible messages via \texttt{rospymoos}.

\subsection{Low-level communications}
Within the ROS network running on the NVIDIA Jetson TX2 onboard each drone, the local planner calculates a nearby ``local waypoint'' for use as a setpoint for the PX4Cube. 
This local waypoint is communicated to the PX4 via the \texttt{mavros}\footnote{https://github.com/mavlink/mavros} ROS node, which is translated into a MAVLINK message using \texttt{mavlink\_router}\footnote{https://github.com/mavlink-router/mavlink-router}, and then finally transmitted from the Jetson TX2 to the PX4 over a universal asynchronous receive/transmit (UART) connection. 


\subsection{Ground station}
To facilitate mission monitoring by humans, we modified the popular open-source QGroundControl (QGC) software, which is commonly used to monitor and control hobby drones. 
The concept was to have one computer with QGC per drone to monitor its status, as well as a ``team'' QGC that displays higher-level information about all the drones in the team.
The major change to QGC was the added capability to access information from each drone's onboard ROS network, which is absent from standard QGC. 
Important algorithmic diagnostic information, such as metrics about the ``health'' of mapping and localisation algorithm, local and global planners, detected targets, and more are now displayed on QGC.

Since a ROS subscriber cannot subscribe to topics managed by two different ROS ``masters'' (i.e. on two different drones) at once, a ROS node was implemented to relay ROS messages to QGC upon receiving an HTTP GET request from QGC. 
This allows each QGC to access ROS information from multiple drones, and for each drone to serve multiple QGC instances on the various ground stations.

\begin{figure}
    \centering
    \includegraphics[width=\linewidth]{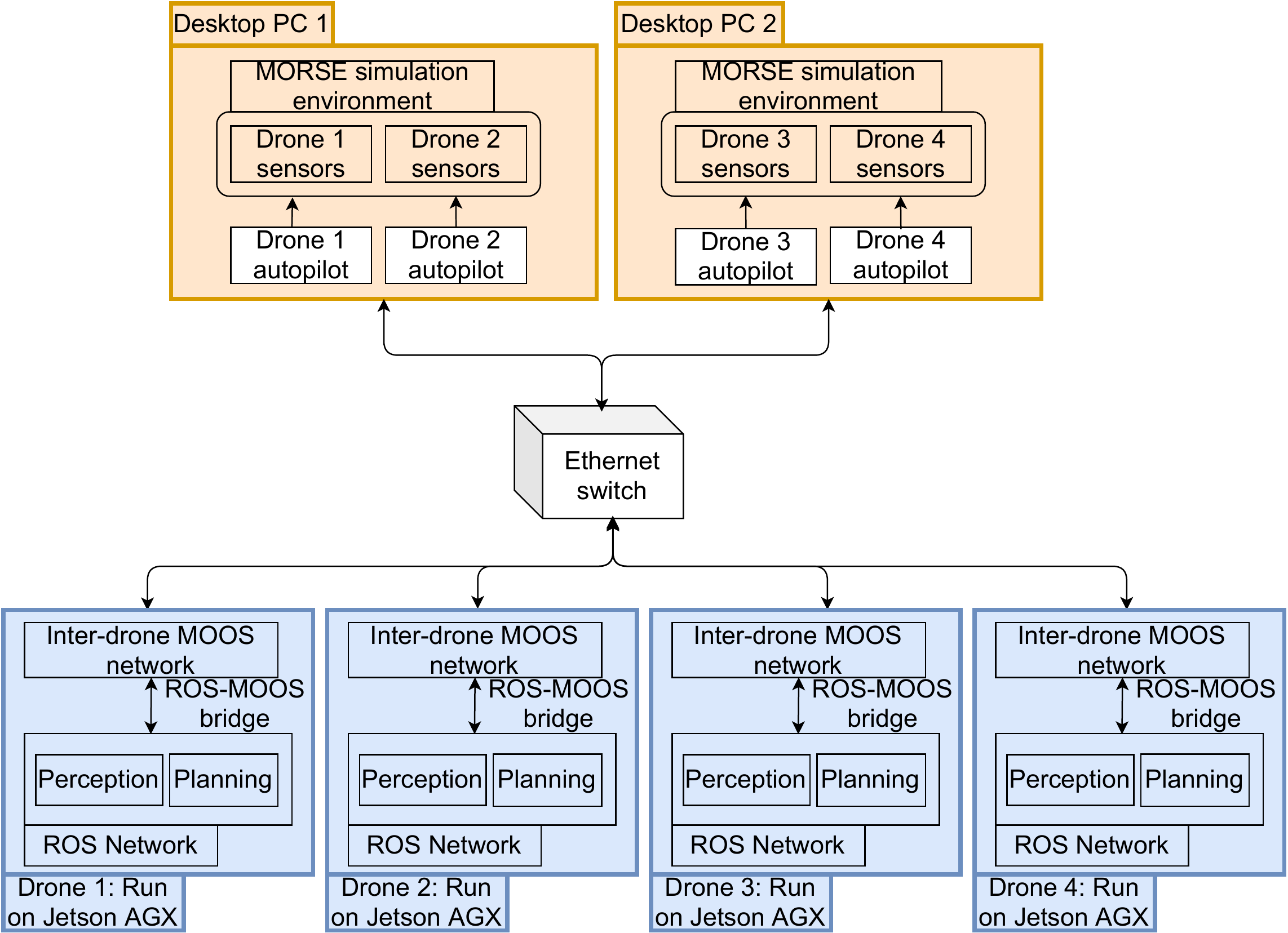}
    \caption{Simulation architecture diagram.}
    \label{fig:sim_architecture}
\end{figure}

\begin{figure*}[t!]
\centering
    \subfloat[]{\includegraphics[height=0.243\textwidth]{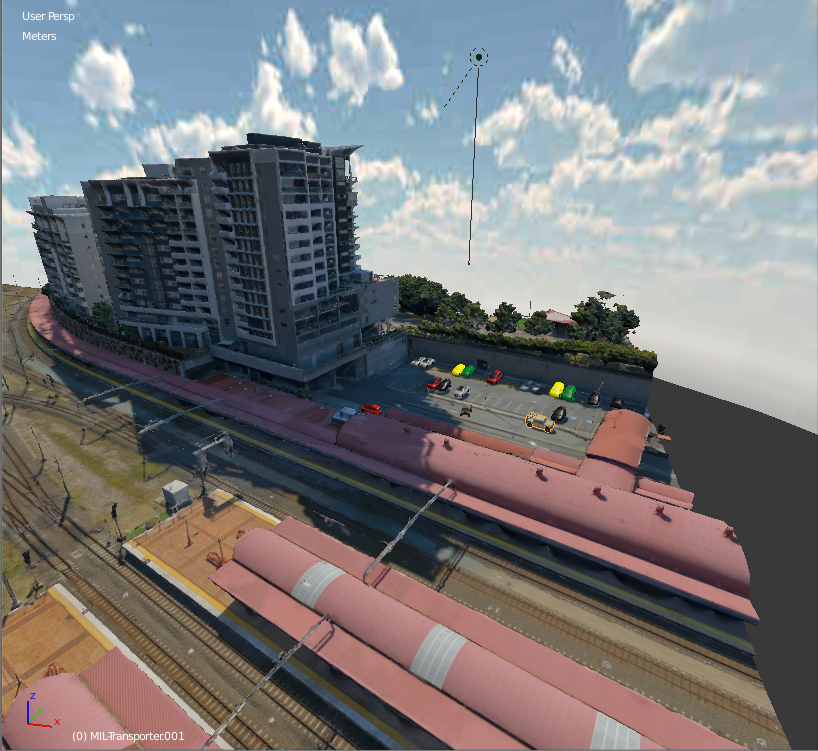}\label{fig:brisbane_env}}    
    \subfloat[]{\includegraphics[width=0.24\textwidth]{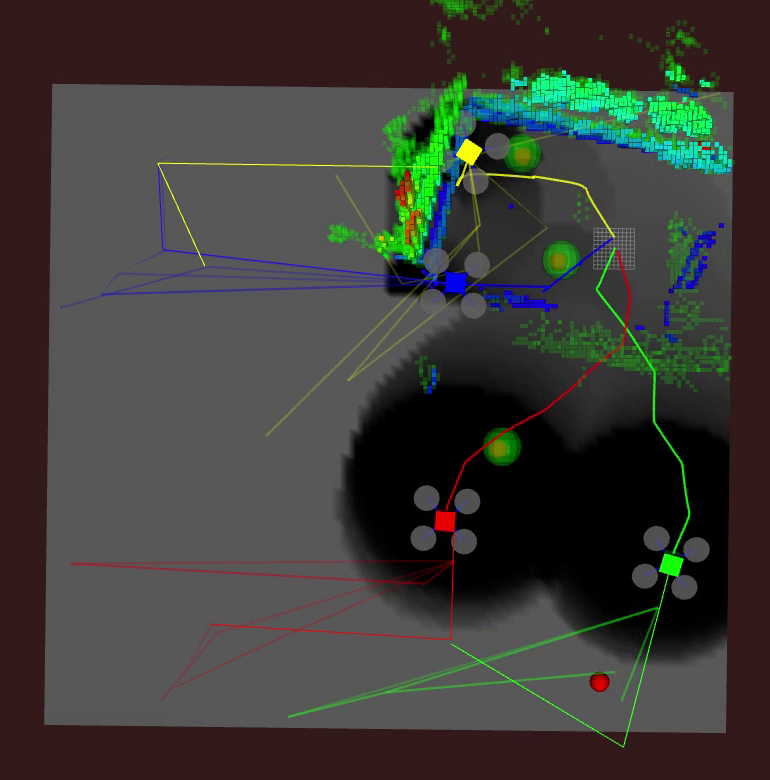}\label{fig:brisbane_2}} 
    \subfloat[]{\includegraphics[width=0.24\textwidth]{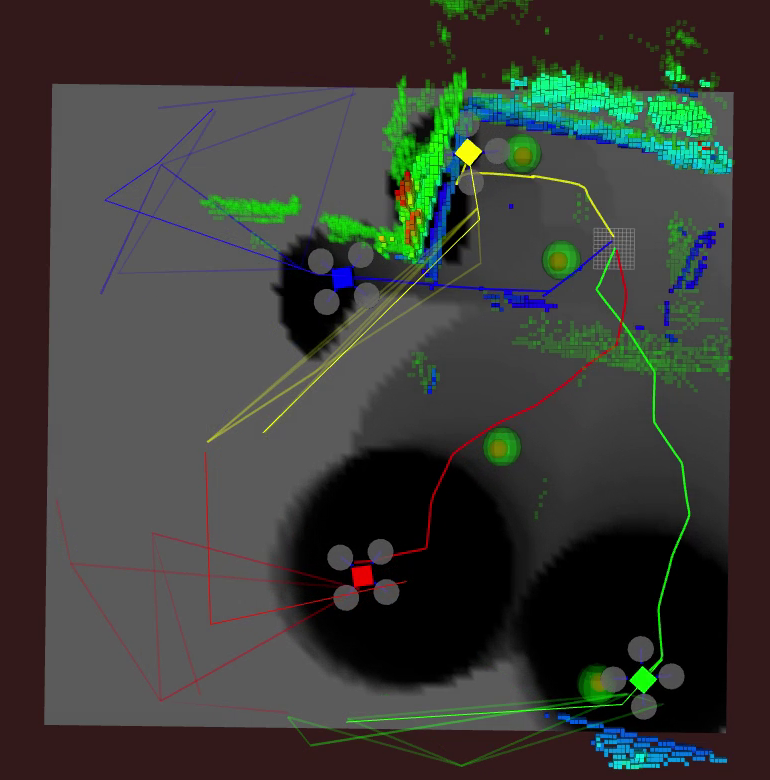}\label{fig:brisbane_3}} 
    \subfloat[]{\includegraphics[width=0.24\textwidth]{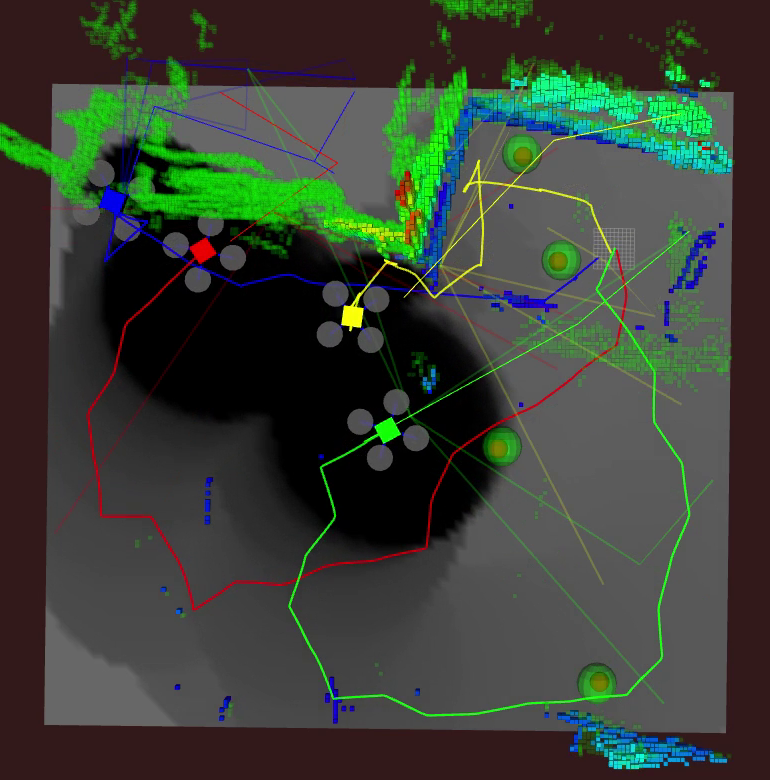}\label{fig:brisbane_4}} \\
    \subfloat[]{\includegraphics[height=0.243\textwidth]{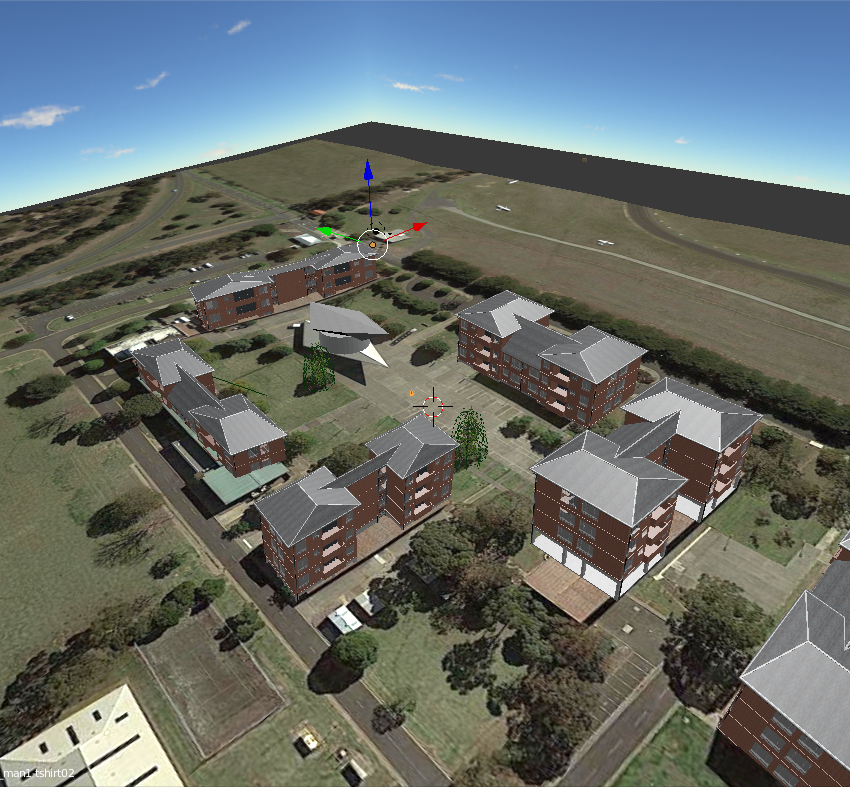}\label{fig:ptcook_env}} 
    \subfloat[]{\includegraphics[width=0.24\textwidth]{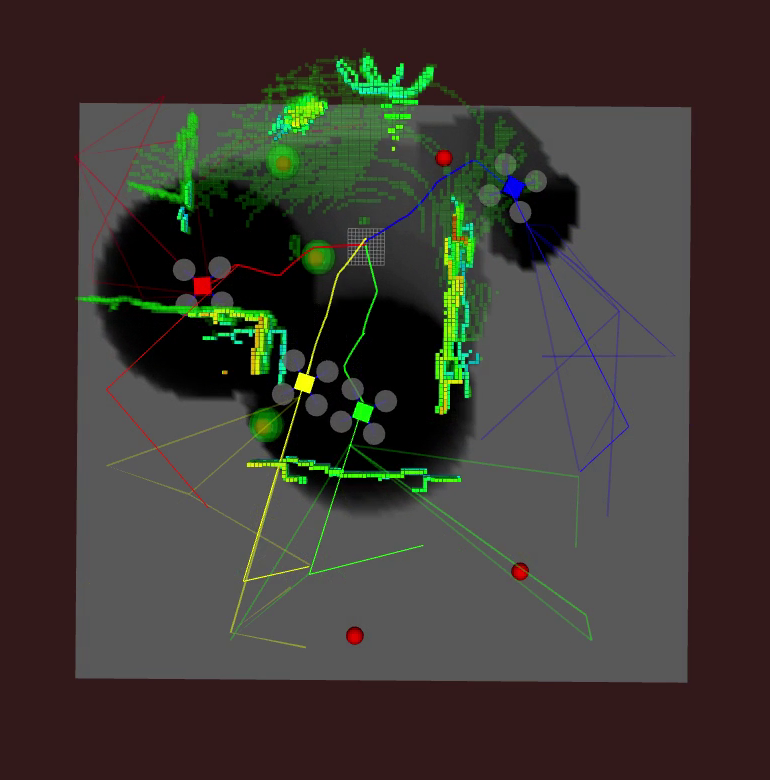}\label{fig:ptcook_2}} 
    \subfloat[]{\includegraphics[width=0.24\textwidth]{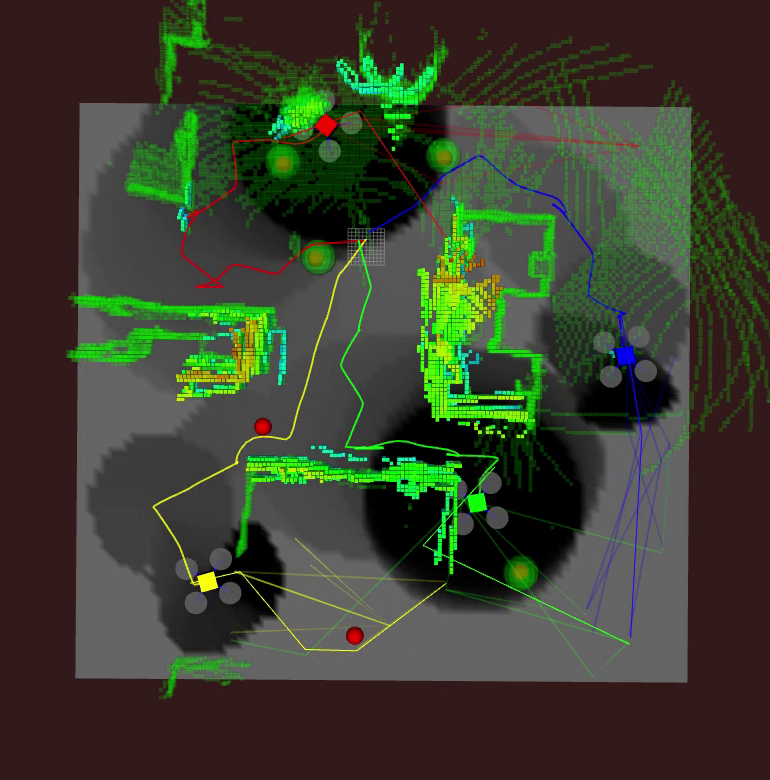}\label{fig:ptcook_3}} 
    \subfloat[]{\includegraphics[width=0.24\textwidth]{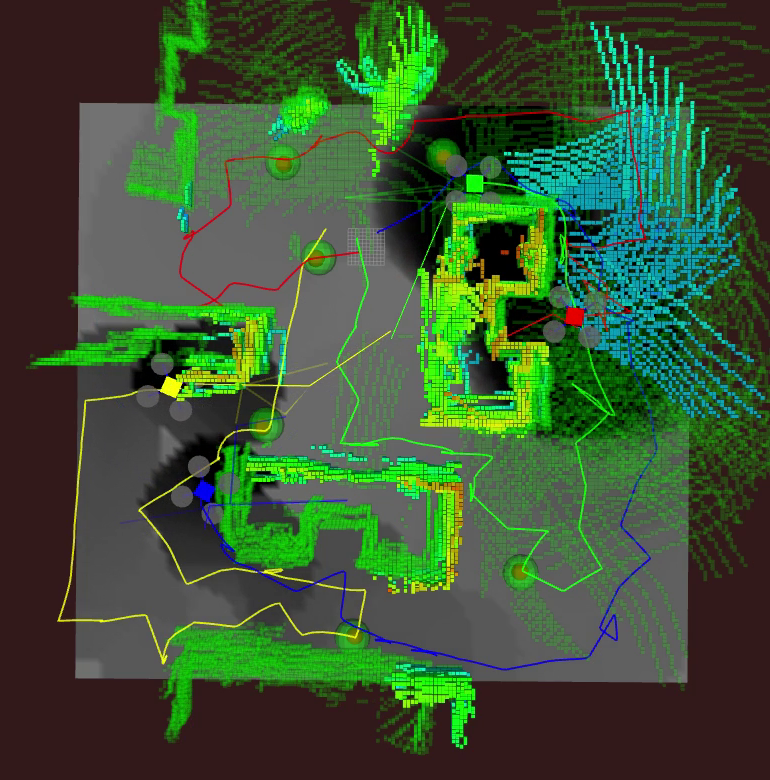}\label{fig:ptcook_4}}
\caption{Simulation results in the Brisbane (a) and Point Cook (e) environments. Time increases to the right. Red spheres are targets yet to be confirmed. They turn green after confirmation. The grey colourmap shows the robot's belief over target. The green structures are the result of decentralised mapping.}
\label{fig:brisbane_rviz}
\vspace{-1ex}
\end{figure*}

\section{Simulation Experiment} \label{sec:results}
Due to COVID-19 restrictions, which were enacted partway through the project, only a limited number of field trials were performed at the Royal Australian Air Force (RAAF) base in Point Cook, Victoria. After restrictions were in place, experiments were performed in simulation instead.

\subsection{Distributed Simulation Framework}

A distributed simulation framework was set up to perform hardware-in-the loop (HIL) testing after field trials became infeasible.
The perception, mapping, and planning software described in Section \ref{sec:software} was used on four NVIDIA Jetson AGX Xaviers, a more powerful but pin-compatible version of the Jetson TX2 onboard the drones.
The Jetson AGX Xaviers are connected via an ethernet switch to two desktop computers with Core i7 processors and NVIDIA RTX2060 graphics cards. 
These desktops simulated the motion of the four drones using the open-source PX4 simulator \cite{meier_px4} in simulated environments (see Fig. \ref{fig:brisbane_rviz}a and \ref{fig:brisbane_rviz}e) using a custom version of the Modular Open Robotics Simulation Engine (MORSE) \cite{misssys_fork}.
The desktops were also responsible for the computationally expensive task of ray-tracing to simulate RGBD and LIDAR sensor measurements for each of the four drones, facilitated through MORSE and Blender. Fig.~\ref{fig:sim_architecture} shows a graphical representation of the simulation architecture.

\subsection{Simulation environments and demonstrations}

In order to replicate real-world conditions, we simulate realistic phenomena, such as varying the belief map over time, and simulating sensor failures such as incorrect point cloud registration.
Despite this, the simulated system appears to perform well and be robust under these conditions.

We first demonstrate the framework in a simulated environment based on Roma St. Station, Brisbane. 
The environment was created from a 3D model generated by high-altitude photogrammetry. 

The results are shown in Figs.~\ref{fig:brisbane_2}-\ref{fig:brisbane_4}.
The team initially starts from the parking lot area.
A nearby target is found upon the start of operation, and is visually confirmed by the blue RGBD drone.
The yellow RGBD drone investigates the corners of parking lots, while the green and red LIDAR drones starts exploring the train station area. 
The blue UAV starts investigating the areas around the building (Fig.~\ref{fig:brisbane_2}). 
As the mission progresses, the green and red UAVs split to south-western and southern parts of the environment. 
In doing so, they each discover one target, and obtain visual confirmation (Fig.~\ref{fig:brisbane_3}), so that all targets are visually confirmed. 

We emphasise that these patterns of ‘allocation’ is a decentralised, emergent behaviour that arises from the coordination between drones, as opposed to centralised task allocation. 
This demonstrates that the Dec-MCTS algorithm is capable of effectively coordinating the scout-task team. 

We also test the robustness of the algorithm against perception failures. 
For this purpose, we use a simulated environment based on the RAAF Williams, Point Cook Base (the ‘Point Cook environment’), depicted in Fig.~\ref{fig:ptcook_env}. 
The environment was prepared by modelling the buildings on top of a satellite image. 

The results are shown in Figs.~\ref{fig:ptcook_2}-\ref{fig:ptcook_4}. 
The drones initially start from the centre of the environment. 
Based on the visible obstacles, the red and blue drones are split to the western and eastern parts of the environment respectively~(Fig.~\ref{fig:ptcook_2}).
Yellow and green drones are allocated to the southern parts of the environment (Fig.~\ref{fig:ptcook_3}). 
Throughout this process, the blue RGBD drone passes over a target but no confirmation is generated~(Fig.~\ref{fig:ptcook_2}). 
The same happens for the target confirmed by the yellow RGBD drone in Fig.~\ref{fig:ptcook_3}. 
Despite this, the target missed by the blue drone is revalidated by the red LIDAR drone in Fig.~\ref{fig:ptcook_3}, and the one missed by the yellow drone is revalidated by the blue drone in Fig.~\ref{fig:ptcook_3}. 
These behaviours demosntrate that the proposed framework can handle perceptual failures that are likely in practice. 

\section{Hardware Experiments} \label{sec:hardware}
\begin{figure}[t]
    \centering
    \includegraphics[width = 0.99\linewidth]{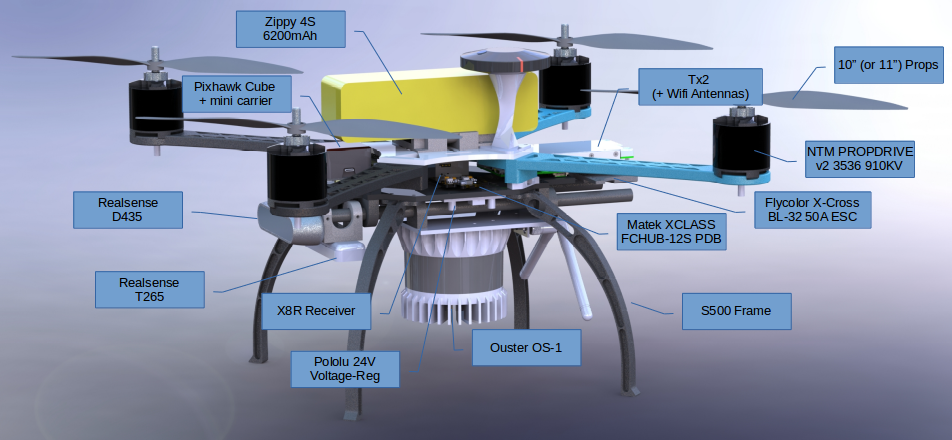}
    \caption{A 3D render of the UAS platform with LIDAR}
    \label{fig:drone}
\end{figure}

\begin{figure}[t]
    \centering
    \includegraphics[width=\linewidth]{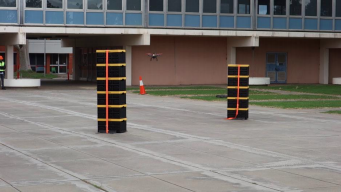}
    \caption{Obstacle avoidance test at the RAAF Williams in Victoria, Australia.}
    \label{fig:hardware_trial}
\end{figure}

The core unit of the multi-robot system is an individual drone (Fig. \ref{fig:drone}). 
The team of drones consists of two types: drones that are task drones, and drones that are both scout and task drones. 
In this team, there are no drones that are ``scout-only'' drones.
All drones share a common airframe and propulsion system, and differ only in their sensor payloads.
The specification is summarised in Table~\ref{tab:spec}. 
In total, the team consists of 4 drones.
Fig. \ref{fig:hardware_trial} shows an image taken during a field trial.

\begin{table}[]
    \centering
    \caption{Specification of the drone platforms}
    \begin{tabular}{r|l}
        Airframe & S500  \\
        Size & 500mm wheelbase \\
        Maximum take-off weight & 1.8 kg \\
        Motors & 4x NTM PROPDRIVE v2 3536 910kV \\
        Propellers & 10-inch x 4.7 \\
        ESC & 4x Flycolor X-cross BL-32 50A \\
        PDB & Matek FCHUB-125 PDB \\
        Battery &  Zippy 4S LiPo, 40C, 6200mAh, 0.59 kg\\
        FCS & Pixhawk Cube \\
        GNSS & Here2 GPS 
    \end{tabular}
    \label{tab:spec}
\end{table}

\subsection{Electronics, flight control hardware}
Low-level control (e.g. attitude control) of each drone is handled by a PX4Cube flight control unit.
Additionally, an NVIDIA Jetson TX2 is onboard each drone to perform higher-level computing tasks, including target recognition, mapping, localisation, planning, and also a ``local planner'' which gives commands to the PX4 flight computer. 

\subsection{Lessons learned from field trials}
Because of the restrictions, full-system field tests of the hardware were not performed. 
However, the field tests that were performed were invaluable in improving and debugging the system.
We describe some lessons learned here.

During field tests, it was discovered that the RealSense D435 camera interfered with the GPS sensor module.
When the D435 plugged into the Jetson TX2 computer onboard the drone, GPS connection would be intermittently lost, or the wrong number of satellites would be reported.
Even when the connection seemed stable, the measurements from the GPS sensor module accumulated unacceptably large drift during test flights, with over 10m of drift in under 5 minutes. 
Since the interference was believed to be electromagnetic, this was remedied by changing the distance between the GPS module and the RealSense D435.
The implemented solution was to be to 3D-print a taller ``mast'' for the GPS module to be raised about 10cm above the body of the drone, which was successful in recovering good GPS sensor performance.
The shorter mast is shown in Fig. \ref{fig:drone}.

While the T265 camera performs SLAM as opposed to simply visual odometry (i.e. it attempts to detect loop closures), in practice this added functionality caused undesirable behavior. 
When flying the drones at low altitude over grass for extended periods of time, impossibly large jumps were detected from the T265 odometry output. 
This was caused by different patches of grass being identified as the same patch, resulting in incorrect loop closures and severe mapping and localisation failure.
This issue affected only the task-only drones, which did not have LIDAR information. 
The problem was ameliorated by detecting the onset of this localisation failure in order to land the drone safely.

\section{Conclusion}\label{sec:conclusion}
Teams of heterogeneous robots have the potential for flexible, reliable, and scalable performance of many useful tasks. 
In this paper we presented an overview of the system in which a heterogeneous multi-drone team was used to search an unknown environment for targets. 
The project is still in progress; as such, we report on encouraging but preliminary results, as well as initial lessons learned in practical implementation of the multi-drone system.

As COVID-19 restrictions ease, we plan to perform further hardware testing to better test the performance of the system in the field.
Hardware testing in other environments (e.g. rural environments) is also an interesting avenue of research; it is likely algorithmic improvements would be required in unstructured terrain.
Looking further ahead, other robots with different specializations (e.g. ground robots with heavier payloads) could be incorporated as well, allowing improved performance and flexibility.

\section*{Acknowledgments}
We wish to thank David Battle, Christian Bruwer, Katrin Schmid, Kwun Yiu Cadmus To, and Rajanish Calisa for their contributions to the software implementation.

\balance
\bibliographystyle{IEEEtran}
\bibliography{references}
\end{document}